\documentclass[11pt,a4paper]{article}
\usepackage{authblk}
\usepackage[hyperref]{ranlp2023}
\usepackage{times}
\usepackage{latexsym}

\usepackage[utf8]{inputenc}

\usepackage{todonotes}

\usepackage{microtype}

\aclfinalcopy 
\def\aclpaperid{***} 

\usepackage{xspace}
\usepackage{cleveref}
\usepackage{multirow}
\usepackage{soul}

\newcommand{\codelink}{\url{https://github.com/kamruzzaman15/BanMANI}}

\newcommand\banmani{\texttt{BanMANI}\xspace}

 \title{
 \banmani:
 A Dataset to Identify Manipulated Social Media News in Bangla}

 \author[1]{\bf{Mahammed Kamruzzaman}}
 \author[2]{\bf{Md. Minul Islam Shovon}}
 \author[3]{\bf{Gene Louis Kim}}
 \affil[1,3]{
   University of South Florida \\
   Tampa, FL, USA 33620}
 \affil[2]{
   Rajshahi University of Engineering \& Technology \\
   Rajshahi-6204, Bangladesh}
 \affil[1,3]{\texttt{\{kamruzzaman1, genekim\}@usf.edu}}
 \affil[2]{\texttt{mainulislam588@gmail.com}}

\date{}

\begin{document}
\maketitle
\begin{abstract}
Initial work has been done to address fake news detection and misrepresentation of news in the Bengali language. However, no work in Bengali yet addresses the identification of specific claims in social media news that falsely manipulates a related news article. At this point, this problem has been tackled in English and a few other languages, but not in the Bengali language. In this paper, we curate a dataset of social media content labeled with information manipulation relative to reference articles, called \banmani. The dataset collection method we describe works around the limitations of the available NLP tools in Bangla. We expect these techniques will carry over to building similar datasets in other low-resource languages. 
\banmani forms the basis both for evaluating the capabilities of existing NLP systems and for training or fine-tuning new models specifically on this task. 
In our analysis, we find that this task challenges current LLMs both under zero-shot and fine-tuned settings.\footnote{Our dataset is available at \codelink.}
\end{abstract}

\section{Introduction}
\label{sec:introduction}

\begin{figure}[t]
\centering
\includegraphics[width=1.0\linewidth]{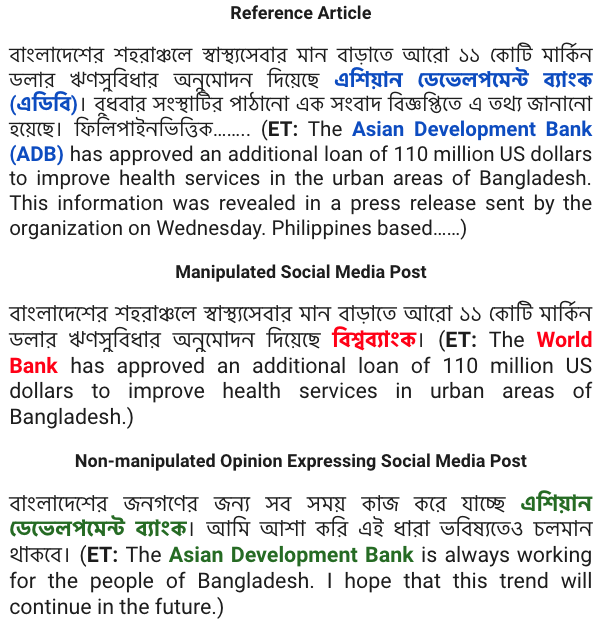}
\caption{
Example of manipulated and non-manipulated social media post with the corresponding reference article. \textbf{ET} denotes the English Translation of the given Bangla sentences. In the given example, the Asian Development Bank (highlighted in blue color) is incorrectly referred to as the World Bank (highlighted in red color) in the manipulated post. In the non-manipulated opinion-expressing post, the Asian Development Bank (highlighted in green color) is correctly referred to.}
\label{fig:example}
\end{figure}
 
Misinformation is an increasingly pressing concern in the current social and political landscape where information frequently spreads through social media platforms with few constraints to reflect the information in reliable sources. This is further exacerbated by the presence of ``bots'' made by malicious actors that are designed to artificially spread ideas that distort reality~\cite{Ferrara_2020,lei-etal-2023-bic}. In order to mitigate this issue, considerable work has been done to identify fake articles~\cite{shu-etal-2020-fakenewsnet}, verifying scientific and encyclopedia claims~\cite{wadden-etal-2020-fact,thorne-etal-2018-fever}, and identifying claims on social media that distort news from trusted sources~\cite{huang2023manitweet}. However, most such work is limited to only English.


Bangla, with the fifth-most L1 speakers worldwide, at 233.7 million\footnote{\url{https://en.wikipedia.org/wiki/List_of_languages_by_total_number_of_speakers}}, only has prior work in detecting fake articles~\cite{hossain-etal-2020-banfakenews}. More work in this direction is needed for Bangla on social media platforms, as demonstrated by the 2012 Ramu incident. In the Ramu incident, a Facebook post from a fake account led to the destruction of a Buddhist temple and dozens of houses in Bangladesh by an angry mob of almost 25,000 people~\cite{inam-manik-2012-hazy}. In this vein, we construct a dataset of news-related social media content for identifying news manipulation in social media, \banmani. This dataset is the comparable Bangla counterpart to the ManiTweet dataset~\cite{huang2023manitweet} in English. \Cref{fig:example} shows an example of a reference news article alongside both a manipulated and a non-manipulated social media post.

This paper's contributions are the following.
\begin{itemize}
    \item We construct a publicly available Bangla dataset of 800 news-related social media items that are annotated as manipulated or not relative to 500 reference news articles.

    \item We present a semi-automatic method for generating such a dataset, which allows scalable dataset collection using annotators efficiently for languages with few available NLP tools.

    \item We demonstrate that current SOTA LLMs struggle on this task, both in zero-shot and fine-tuned settings. 
\end{itemize}

\section{Related Work}
This paper is most closely related to fact-checking and fake news detection 
tasks. While much work in this direction has been done in English, 
\citet{hossain-etal-2020-banfakenews} only recently started work in this domain in Bangla by releasing a dataset for fake news detection. 

In English, \citet{huang2023manitweet} released a dataset for identifying news manipulation in Tweets. 
In order to supplement fully-human data, they used a semi-automatic approach of generating Tweets using ChatGPT and using human annotators to validate and label the results. 
They found that ChatGPT and Vicuna failed to solve this new task, even after fine-tuning. 
In their work, they used \texttt{FakeNewsNet}~\cite{shu2019fakenewsnet} dataset to seed their reference articles.




Fact-checking tasks closely resemble our task in that claims must be compared against reference evidence, such as in the FEVER~\cite{thorne-etal-2018-fever} and SCIFACT~\cite{wadden-etal-2020-fact} datasets. Techniques for this kind of fact-checking work often use a retrieval module that pulls relevant data from the supplied candidate pool. The degree of consistency between a piece of evidence and the input claim is then evaluated using a reasoning component~\cite{pradeep-etal-2021-scientific}.

While our task compares text against a reference article, models must be able to separate social media news related to the reference article from those that only convey opinions to ensure the successful completion of our task. This is the key difference between these (i.e., fact-checking 
and fake news detection) and our work. 

\section{
Task Definition
}
Our goal is to identify whether a news-related \textit{social media item} 
(a post or a comment) is manipulated. If the social media item is manipulated then furthermore to determine what particular information is being manipulated relative to a related reliable reference article.
We divide this task into three parts. 
\paragraph{Subtask 1.}
First, we identify whether a particular social media item is manipulated. This part is a binary classification task and we consider an item as manipulated if there is at least one manipulated excerpt.

\paragraph{Subtask 2.}
Second, if a social media item is classified as manipulated then we need to identify which particular excerpt is manipulated. The task then is to identify the excerpt of the social media item which is not consistent with the original reference news article. In our dataset, we refer to any manipulated or newly introduced span
as an \textit{altered excerpt}.

\paragraph{Subtask 3.}
The third subtask is to identify the part of the original news article which is manipulated in the social media item. In our dataset, we define the information being manipulated as \textit{original excerpt}. Models must produce an empty string or ``none'' as the output when the \textit{altered excerpt} is inserted without modifying any \textit{original excerpts}.

\section{\banmani: Dataset Creation 
}

\begin{figure*}
    \includegraphics[width=1.0\linewidth]{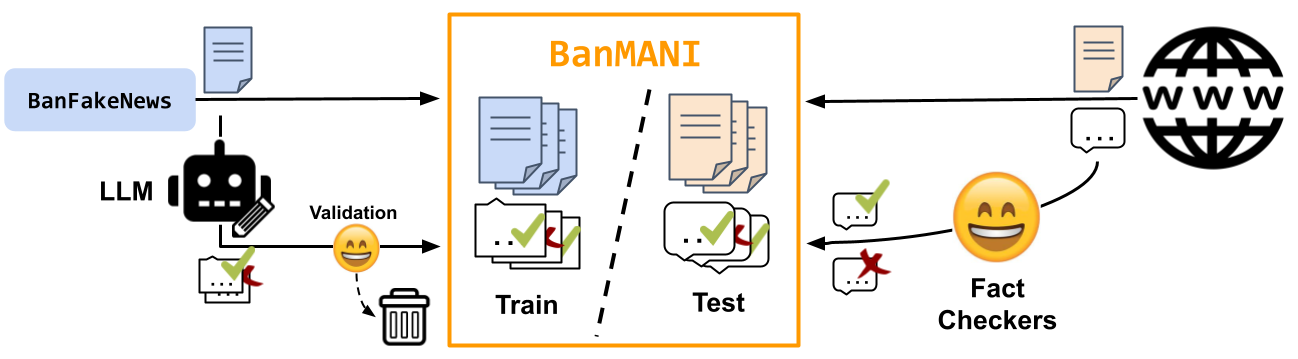}
    \caption{A diagram of the dataset collection procedure. The left side shows the semi-automatic data collection procedure for the training set, seeded by the BanFakeNews dataset~(\Cref{ssec:social-media-item-generation}). The right side shows the collection of human-fact-checked items for the test set~(\Cref{ssec:test-data-collection}).}
    \label{fig:dataset-collection}
\end{figure*}

We confined our test data collection to Facebook
since this platform is more commonly used by Bangla speakers compared to typically studied media platforms for English speakers (e.g., Twitter, Instagram, etc.).\footnote{According to StatCounter.com~(\url{https://gs.statcounter.com/}), Twitter held 22.01\% of thie social media market share in the US in June 2023, but only 1.41\% in Bangladesh. On the other hand, Facebook held 48.2\% of the market share in the US but 78.84\% in Bangladesh.} We have created a dataset that contains 800 news-related social media items with 500 associated news articles. Our dataset contains 530 manipulated items and 270 non-manipulated items. The breakdown of our dataset is shown in \Cref{tab:sta}.

\begin{table*}
\begin{center}
  \begin{tabular}{|l|l|l|l|l|l|l|}
    \hline
    \multirow{2}{*}{Split} &
      \multicolumn{2}{c}{Manipulated} &
      \multicolumn{2}{c|}{Non-manipulated} & \\
      
    & Post & Comment & Post & Comment  & Total \\
    \hline
    Train & 370 & 100 & 130 & 50 & 650\\
    \hline
    Test & 40 & 20 & 60 & 30 & 150 \\
    \hline
  \end{tabular}
\end{center}
\caption{\label{tab:sta} \banmani Dataset Statistics}
\end{table*}

\subsection{Bangla-specific Challenges}
The task of constructing a Bangla version of the \textsc{ManiTweet} dataset is complicated by several factors. First and foremost, the availability and efficacy of NLP tools in Bangla are much more limited than in English. This means that some reliably automated steps in the English data collection process may be impossible or unreliable in Bangla. In addition to this, a Bangla version of \texttt{FakeNewsNet}~\cite{shu-etal-2020-fakenewsnet}, the dataset that \citet{huang2023manitweet} use as a basis for their \textsc{ManiTweet} dataset, does not exist. \texttt{FakeNewsNet} contains news articles with associated Twitter data which can be directly annotated with any identifiable manipulation. In our dataset construction process, we must identify news articles and corresponding social media posts ourselves since no such seed dataset exists in Bangla.

\subsection {Source of News Articles}
We collected our news article from \texttt{BanFakeNews} 
\cite{hossain-etal-2020-banfakenews}, a dataset for Bangla fake news detection. From that dataset, we selected 6 domains where we expect the most social media manipulation to occur: National, International, Politics, Entertainment, Crime, and Finance. 
From those categories, we selected 2.3k seed news articles, which were used to generate manipulated and non-manipulated social media news.
We furthermore upsampled the Politics and Entertainment domains as these were singled out in \citeauthor{huang2023manitweet}'s~(\citeyear{huang2023manitweet}) analysis.
For more details on the initial data selection, see \Cref{app:dataset}.

\subsection{
Social Media Item Generation}
\label{ssec:social-media-item-generation}

No suitable dataset of social media items with corresponding news articles exists in Bangla. In order to efficiently use our limited annotator resources, we deploy a semi-automated data collection process using ChatGPT\footnote{GPT-3.5-turbo}. We use ChatGPT to generate both manipulated and non-manipulated social media items from a seed news article, which is then validated by human annotators.


\subsubsection{Collection of Substitutable Sets}
In order to generate manipulated social media items using ChatGPT, we first must identify plausible but incorrect substitutions that can be made in social media items. We collect such possible substitutions through a named entity recognition~(NER) tagger. This mirrors the procedure used by \citet{huang2023manitweet}.
We collect news-relevant substitutable sets by running a Bangla NER tagger on 2,300 news articles from the \texttt{BanFakeNews}. We consider any two entities with the same NER label as substitutable with each other. We collected all \texttt{PERSON}, \texttt{ORGANIZATION}, and \texttt{LOCATION} named entities from the NER results, following the NER label choices used by \citet{huang2023manitweet}.

Based on preliminary experimentation of available Bangla NER systems, we found mBERT-Bengali-NER\footnote{\url{https://huggingface.co/sagorsarker/mbert-bengali-ner}},
a BERT-based multilingual Bengali NER system, to perform the best in our use case. Due to the high error rate of Bangla NER taggers, we perform a human filtering step to remove mistakes in the automatic NER labeling. Details of this step are provided in \Cref{app:NER}.

We supplement the automatically collected entity sets with manually constructed sets of common entity substitutions that were identified in the data construction process.
For example, some people write 
\textbf{Asian Development Bank} in their post, when the original news article contains 
\textbf{World Bank}. The same interchange also happens for 
\textbf{Bangladesh Bank} and 
\textbf{Asian Infrastructure Investment Bank}. So we created a substitutable subset inside the \texttt{ORGANIZATION} entity label that contains these four together 
Members of these hand-curated sets can similarly be substituted with each other to create manipulated news.




\subsubsection{Item Generating Prompts}
\label{ssec:item-generation}

\begin{figure}[t]
\centering
\includegraphics[width=1.0\linewidth]{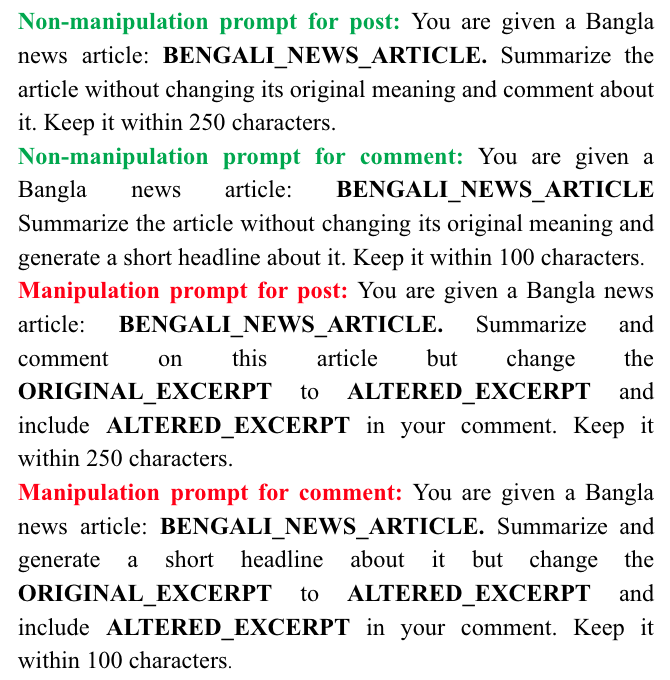}
\captionof{figure}{Prompt templates for social media item generation. Here, the ``ALTERED EXCERPT'' and ``ORIGINAL EXCERPT'' bear the same meaning as described in Subtask 2 and 3 respectively. }
\label{fig:prmpt}
\end{figure}

We use the \texttt{content} attribute of the news articles from the \texttt{BanFakeNews} to create the social media posts and \texttt{headline} attribute of the news articles to create comments. Since comments are generally shorter than posts, we use a different approach to generate them. Social media item generation prompt templates are given in \Cref{fig:prmpt}.





After generating the manipulated and non-manipulated social media items using ChatGPT, we assign human annotators to validate the generated data. The total number of generating manipulated and non-manipulated items using ChatGPT are 2.3k. The generated social media items from ChatGPT are not always coherent or related to the seed news articles. So the human annotators discarded 1.65k generated data during the validation stage. We use the remaining social media items generated by ChatGPT as our training data. In this project, graduate and undergraduate students are working as human annotators. The inter-annotator agreement between the involved annotators is 92.2\% per Cohen’s kappa \cite{cohen1960coefficient}. The detailed data annotation process, including screenshots of the annotation interfaces, is available in \Cref{app:data-annotation}.

\subsection {Test Data Collection} 
\label{ssec:test-data-collection}

We collected 150 human-generated social media items for our test set. These items were collected manually from Facebook using two distinct strategies. In the first strategy, items were sourced from media and news company pages on Facebook, such as Prothom Alo.\footnote{\url{https://www.facebook.com/DailyProthomAlo}} From these pages, we collected posts that shared a news article with accompanying post text that add commentary as well as comments from the comment sections under news articles on the page. In the second strategy, we collected posts from pages such as BD FactCheck \footnote{\url{https://www.facebook.com/bdfactcheck}} and Rumor Scanner \footnote{\url{https://www.facebook.com/RumorScanner}} which specialize in identifying fake news published on other platforms. 

\begin{table*}[t]
\begin{center}
\centering
\begin{tabular}{llll}
\hline
\textbf{Metric} & \textbf{Subtask 1} & \textbf{Subtask 2} & \textbf{Subtask 3}\\
\hline
F1 & 57.02 & \texttt{--} & \texttt{--} \\
EM & \texttt{--} & 8.2 & 12.3 \\
RL (r, p, f) & \texttt{--} & (79.72, 36.06, 46.83) & (64.78, 41.04, 49.94)\\
\hline
\end{tabular}
\end{center}
\caption{\label{tab:zeroshot}Evaluation results of ChatGPT with Zero-shot. Here, EM denotes Exact Match, RL denotes ROUGE-L, which is broken down into (r, p, f) denoting recall, precision, and F1 score respectively.}
\end{table*}

\begin{table*}
\begin{center}
\centering
\begin{tabular}{llll}
\hline
\textbf{Metric} & \textbf{Subtask 1} & \textbf{Subtask 2} & \textbf{Subtask 3}\\
\hline
F1 & 65.77 & \texttt{--} & \texttt{--} \\
EM & \texttt{--} & 11.9 & 13.34 \\
RL (r, p, f) & \texttt{--} & (61.95, 69.26, 64.75) & (63.65, 50.74, 56.46)\\
\hline
\end{tabular}
\end{center}
\caption{\label{tab:fine-tuned}Evaluation result of our fine-tuned GPT-3 model. The table labeling conventions match those of \Cref{tab:zeroshot}.}
\end{table*}

\section{Exploratory Data Analysis}
From \Cref{tab:testdata}, we see most of the manipulated news is political. Some people spread manipulated news on social media to influence public opinion, promote a particular political party, etc., and these might be the reasons behind the manipulated political news. Also, we notice that national and international news are manipulated in a bigger amount. Sites and pages with low trustworthiness are most likely to spread manipulated news. The followers of those sites and pages are most likely unaware of the fact and accidentally post manipulated news. 

\begin{table}
\begin{center}
\centering
\begin{tabular}{ll}
\hline
\textbf{Domain} & \textbf{Manipulated Articles}\\
\hline
National & 16 \\
International & 14 \\
Politics & 19 \\
Entertainment & 5 \\
Crime & 1  \\
Finance & 5 \\
\hline
\end{tabular}
\end{center}
\caption{\label{tab:testdata} Manipulated News Articles in Test Data}
\end{table}

\section{Experimental Setup}


\subsection{Models}
\paragraph{Zero-shot ChatGPT.}
We use ChatGPT for the zero-shot setting experiments. For details prompt about the zero-shot experiment, see \Cref{app:zeroshot}. 

\paragraph{Fine-tuned.}
Fine-tuning allows the user to get more out of the available models through provided API. 
As a result, it can achieve higher quality results than traditional prompt design, train on more examples beyond the limit of traditional prompt, and saves token due to shorter prompts. Fine-tuning improves on few-shot learning by training on much more examples that can fit in a prompt. Which lets you achieve better results in fine-tuned tasks. 
In general, fine-tuning involves preparing and uploading training data, training the new fine-tuned model with prepared data, and using the fine-tuned model. For our work, we used GPT-3~\cite{brown2020language} \texttt{ada}\footnote{\url{https://platform.openai.com/docs/models}} as our base model due to the unavailability of fine-tuning for the latest models. Also, \texttt{ada} is capable of handling simple tasks and is the fastest model in the GPT-3 series. We used a prompt-completion format for our training data and later fine-tuned our model with this data, resulting in competitive outputs.

\subsection{Evaluation Metrics}
For subtask 1, we use F1 score as this is simply a classification task. Since subtasks 2 and 3 involve span extraction, we use Exact Match (EM) and ROUGE-L (RL).

\section{Results \& Analysis}
\label{sec:results+analysis}

The result of the zero-shot ChatGPT and our fine-tuned model is presented in \Cref{tab:zeroshot} and \Cref{tab:fine-tuned} respectively. From \Cref{tab:zeroshot} and \Cref{tab:fine-tuned}, we can see that our fine-tuned model outperforms the zero-shot ChatGPT model for subtask 1, where the F1 score of zero-shot ChatGPT and fine-tuned model is 57.02\% and 65.77\% respectively. In terms of EM, we can see that our fine-tuned model performs better for both subtask 2 and subtask 3. For subtask 2, if we look at the RL value of our fine-tuned model, we can see that the precision of RL is 69.26\%, which is 33.2\% more than the zero-shot model. That is also the case for the F1 score of RL. In the same way, for sub-task 3, we can see that the precision and F1 score of RL outperforms the zero-shot model.

\section{Limitations \& Future Work}
Due to our budget limitation, we were not able to collect a large set of human-written social media items. This means that there exists a gap between the quality of the training and test data; the training set was automatically created, unlike the test data. In the future, we will collect more human-written items from social media to create an entirely human-written training dataset. Our prompts are also purposefully simple, as this was the first step in creating such a dataset. We expect to get qualitative gains in the automatically generated data with more careful prompt engineering. Finally, our experiments were limited to only a single popular LLM for each setting. Expanding the experiments to cover other LLMs, especially open-source LLMs would lead to more robust experimental results and better replicability. We also leave the few-shot method as our future work. 

\section{Conclusion}
\label{sec:conclusion}

In this paper, we presented the \banmani dataset, the semi-automatically constructed dataset of news manipulation in social media. This dataset extends \citeauthor{huang2023manitweet}'s~\citeyear{huang2023manitweet} \textsc{ManiTweet} dataset to Bangla. Our semi-automatic collection process generates social media posts from seed articles using a multilingual LLM and a Bangla NER system. These results are filtered using human annotators for efficient use of annotator time. We find that both zero-shot and fine-tuned LLMs struggle on this dataset, pointing to important directions of future work. Surprisingly, we find that LLMs perform similarly effectively on this dataset when compared to the English variant. We hope that this new resource can help with combating information manipulation in Bangla-speaking social media communities. Furthermore, we believe that the technique laid out here can act as a basis for similar work in other under-served languages in NLP.




\bibliography{anthology,custom}

\begin{thebibliography}{12}
\expandafter\ifx\csname natexlab\endcsname\relax\def\natexlab#1{#1}\fi

\bibitem[{Ahmed and Manik(2012)}]{inam-manik-2012-hazy}
Inam Ahmed and Julfikar~Ali Manik. 2012.
\newblock \href {https://www.thedailystar.net/news-detail-252212} {A hazy
  picture appears}.
\newblock [Online; posted 03-October-2012].

\bibitem[{Brown et~al.(2020)Brown, Mann, Ryder, Subbiah, Kaplan, Dhariwal,
  Neelakantan, Shyam, Sastry, Askell, Agarwal, Herbert-Voss, Krueger, Henighan,
  Child, Ramesh, Ziegler, Wu, Winter, Hesse, Chen, Sigler, Litwin, Gray, Chess,
  Clark, Berner, McCandlish, Radford, Sutskever, and
  Amodei}]{brown2020language}
Tom~B. Brown, Benjamin Mann, Nick Ryder, Melanie Subbiah, Jared Kaplan,
  Prafulla Dhariwal, Arvind Neelakantan, Pranav Shyam, Girish Sastry, Amanda
  Askell, Sandhini Agarwal, Ariel Herbert-Voss, Gretchen Krueger, Tom Henighan,
  Rewon Child, Aditya Ramesh, Daniel~M. Ziegler, Jeffrey Wu, Clemens Winter,
  Christopher Hesse, Mark Chen, Eric Sigler, Mateusz Litwin, Scott Gray,
  Benjamin Chess, Jack Clark, Christopher Berner, Sam McCandlish, Alec Radford,
  Ilya Sutskever, and Dario Amodei. 2020.
\newblock \href {http://arxiv.org/abs/2005.14165} {Language models are few-shot
  learners}.

\bibitem[{Cohen(1960)}]{cohen1960coefficient}
Jacob Cohen. 1960.
\newblock A coefficient of agreement for nominal scales.
\newblock \emph{Educational and psychological measurement}, 20(1):37--46.

\bibitem[{Ferrara(2020)}]{Ferrara_2020}
Emilio Ferrara. 2020.
\newblock \href {https://doi.org/10.5210/fm.v25i6.10633} {What types of
  covid-19 conspiracies are populated by twitter bots?}
\newblock \emph{First Monday}, 25(6).

\bibitem[{Hossain et~al.(2020)Hossain, Rahman, Islam, and
  Kar}]{hossain-etal-2020-banfakenews}
Md~Zobaer Hossain, Md~Ashraful Rahman, Md~Saiful Islam, and Sudipta Kar. 2020.
\newblock \href {https://aclanthology.org/2020.lrec-1.349} {{B}an{F}ake{N}ews:
  A dataset for detecting fake news in {B}angla}.
\newblock In \emph{Proceedings of the Twelfth Language Resources and Evaluation
  Conference}, pages 2862--2871, Marseille, France. European Language Resources
  Association.

\bibitem[{Huang et~al.(2023)Huang, Chan, McKeown, and Ji}]{huang2023manitweet}
Kung-Hsiang Huang, Hou~Pong Chan, Kathleen McKeown, and Heng Ji. 2023.
\newblock \href {http://arxiv.org/abs/2305.14225} {Manitweet: A new benchmark
  for identifying manipulation of news on social media}.

\bibitem[{Lei et~al.(2023)Lei, Wan, Zhang, Feng, Chen, Li, Zheng, and
  Luo}]{lei-etal-2023-bic}
Zhenyu Lei, Herun Wan, Wenqian Zhang, Shangbin Feng, Zilong Chen, Jundong Li,
  Qinghua Zheng, and Minnan Luo. 2023.
\newblock \href {https://aclanthology.org/2023.acl-long.575} {{BIC}: {T}witter
  bot detection with text-graph interaction and semantic consistency}.
\newblock In \emph{Proceedings of the 61st Annual Meeting of the Association
  for Computational Linguistics (Volume 1: Long Papers)}, pages 10326--10340,
  Toronto, Canada. Association for Computational Linguistics.

\bibitem[{Pradeep et~al.(2021)Pradeep, Ma, Nogueira, and
  Lin}]{pradeep-etal-2021-scientific}
Ronak Pradeep, Xueguang Ma, Rodrigo Nogueira, and Jimmy Lin. 2021.
\newblock \href {https://aclanthology.org/2021.louhi-1.11} {Scientific claim
  verification with {V}er{T}5erini}.
\newblock In \emph{Proceedings of the 12th International Workshop on Health
  Text Mining and Information Analysis}, pages 94--103, online. Association for
  Computational Linguistics.

\bibitem[{Shu et~al.(2020)Shu, Mahudeswaran, Wang, Lee, and
  Liu}]{shu-etal-2020-fakenewsnet}
K.~Shu, D.~Mahudeswaran, S.~Wang, D.~Lee, and H.~Liu. 2020.
\newblock {{F}ake{N}ews{N}et: {A} {D}ata {R}epository with {N}ews {C}ontent,
  {S}ocial {C}ontext, and {S}patiotemporal {I}nformation for {S}tudying {F}ake
  {N}ews on {S}ocial {M}edia}.
\newblock \emph{Big Data}, 8(3):171--188.

\bibitem[{Shu et~al.(2019)Shu, Mahudeswaran, Wang, Lee, and
  Liu}]{shu2019fakenewsnet}
Kai Shu, Deepak Mahudeswaran, Suhang Wang, Dongwon Lee, and Huan Liu. 2019.
\newblock \href {http://arxiv.org/abs/1809.01286} {Fakenewsnet: A data
  repository with news content, social context and spatialtemporal information
  for studying fake news on social media}.

\bibitem[{Thorne et~al.(2018)Thorne, Vlachos, Christodoulopoulos, and
  Mittal}]{thorne-etal-2018-fever}
James Thorne, Andreas Vlachos, Christos Christodoulopoulos, and Arpit Mittal.
  2018.
\newblock \href {https://doi.org/10.18653/v1/N18-1074} {{FEVER}: a large-scale
  dataset for fact extraction and {VER}ification}.
\newblock In \emph{Proceedings of the 2018 Conference of the North {A}merican
  Chapter of the Association for Computational Linguistics: Human Language
  Technologies, Volume 1 (Long Papers)}, pages 809--819, New Orleans,
  Louisiana. Association for Computational Linguistics.

\bibitem[{Wadden et~al.(2020)Wadden, Lin, Lo, Wang, van Zuylen, Cohan, and
  Hajishirzi}]{wadden-etal-2020-fact}
David Wadden, Shanchuan Lin, Kyle Lo, Lucy~Lu Wang, Madeleine van Zuylen, Arman
  Cohan, and Hannaneh Hajishirzi. 2020.
\newblock \href {https://doi.org/10.18653/v1/2020.emnlp-main.609} {Fact or
  fiction: Verifying scientific claims}.
\newblock In \emph{Proceedings of the 2020 Conference on Empirical Methods in
  Natural Language Processing (EMNLP)}, pages 7534--7550, Online. Association
  for Computational Linguistics.

\end{thebibliography}
\bibliographystyle{acl_natbib}

\appendix

\section{Initial Data selection Details}
\label{app:dataset}
Initially (before the first round of human validation) we took 2.3k news articles and generated news-related social media items. In \Cref{tab:dataset}, we show the details of each category data. 

\section{NER Annotation Process}
\label{app:NER}
\begin{figure}
\centering
\includegraphics[width=1.0\linewidth]{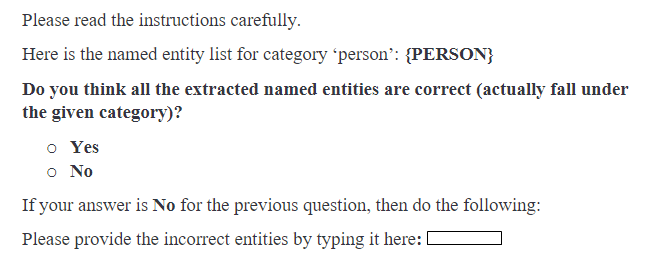}
\captionof{figure}{NER Annotation Interface}
\label{fig:ner}
\end{figure}

Since the performance of the Bangla NER system is not accurate, we need to discard some of the named entities after extracting them. We presented our NER annotation details in \Cref{fig:ner}. In \Cref{fig:ner}, we only show the annotation process for \texttt{PERSON} and we do this for every other category (i.e, \texttt{ORGANIZATION, and LOCATION}). 

\section{Data Annotation Process}
\label{app:data-annotation}
In our research, we perform a two-stage data annotation process for our data. To ensure data quality and consistency, we have selected only those annotators whose mother tongue is Bengali. In this project, all the annotators are graduate and undergraduate students from different institutions. In this project, we have selected a total of 5 students as annotators and kept the data that got at least three annotators' votes.

\begin{figure}
\centering
\includegraphics[width=1.0\linewidth]{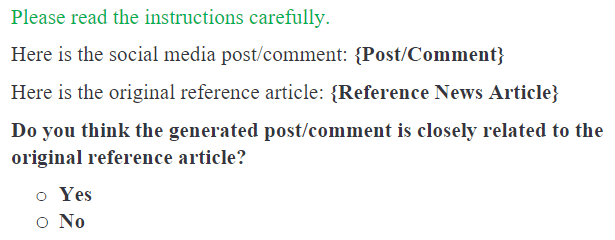}
\captionof{figure}{Stage 1 Annotation.}
\label{fig:stage1}
\end{figure}

\paragraph{Stage 1.} In the first stage, we asked each annotator to read the generated social media items carefully and see whether it makes sense to them or not and this stage is only limited to our train data. We need to introduce  this round because sometimes ChatGPT generates very poor data that doesn't make any sense or totally unrelated to the corresponding news article. Especially the Bangla data generation performance of ChatGPT is bad compared to English. So this round of annotation ensures the generated items are not unrelated to the news topic. \Cref{fig:stage1}  represents the annotation details of stage 1. We only keep those data that receive a `Yes' in stage 1.

\begin{table}
\begin{center}
\centering
\begin{tabular}{ll}
\hline
\textbf{Domain} & \textbf{No. of Articles}\\
\hline
National & 288 \\
International & 288 \\
Politics & 690 \\
Entertainment & 460 \\
Crime & 287  \\
Finance & 287 \\
\hline
\end{tabular}
\end{center}
\caption{\label{tab:dataset} Initially Taken News Articles Based on Each Category}
\end{table}

\begin{figure}
\centering
\includegraphics[width=1.0\linewidth]{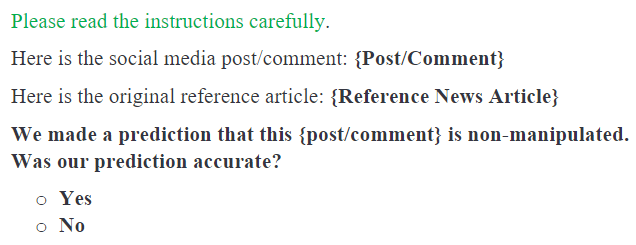}
\captionof{figure}{Stage 2 Annotation for Non-manipulated Social Media Items.}
\label{fig:non-mani}
\end{figure}

\begin{figure}
\centering
\includegraphics[width=1.0\linewidth]{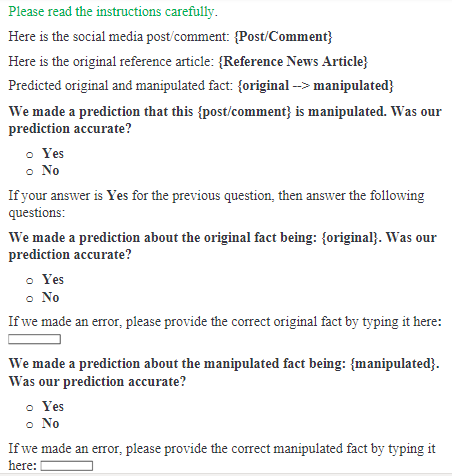}
\captionof{figure}{Stage 2 Annotation for Manipulated Social Media Items.}
\label{fig:mani}
\end{figure}

\paragraph{Stage 2.}  Here in stage two, we annotated our test and train both data based on manipulated and non-manipulated classes. For the non-manipulation class, we follow the instructions pictured in \Cref{fig:non-mani}. The annotation interface for the manipulated class is presented in \Cref{fig:mani}. We keep the data that receive a 'Yes' for non-manipulated class. For the manipulated class, we asked a few more questions for annotators because it is difficult to collect manipulated data from social media. If the answer to the first annotation interface question for manipulated class is 'Yes', then we asked two more questions. The purpose of the latter two questions is that if we classified the manipulated post correctly but accidentally got the altered or original excerpt wrong, then the annotators can give us the accurate excerpt and in this way, we can keep the data. 

\section{Zero-shot Prompt for ChatGPT}
\label{app:zeroshot}
\begin{figure}
\centering
\includegraphics[width=1.0\linewidth]{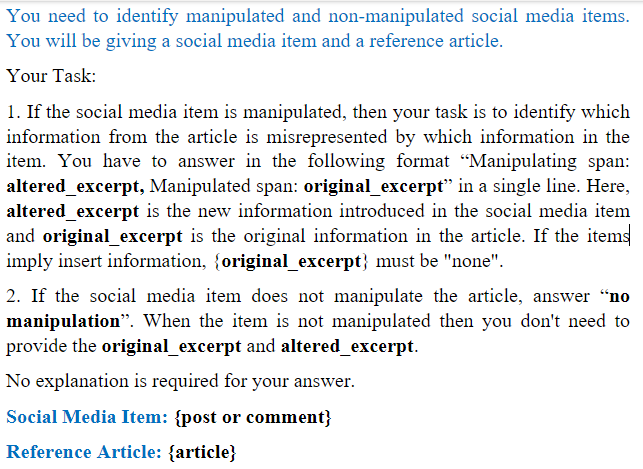}
\captionof{figure}{Zero-shot Prompt}
\label{fig:zeroshot}
\end{figure}

The zero-shot prompt template for the ChatGPT model is shown in \Cref{fig:zeroshot}.

\end{document}